\documentclass{article}
\usepackage{ijcai26}

\usepackage{times}
\usepackage{soul}
\usepackage{url}
\usepackage[hidelinks]{hyperref}
\usepackage[utf8]{inputenc}
\usepackage[small]{caption}
\usepackage{graphicx}
\usepackage{amsmath}
\usepackage{amsthm}
\usepackage{booktabs}
\usepackage{algorithm}
\usepackage{algorithmic}
\usepackage[switch]{lineno}

\usepackage[caption=false]{subfig}
\usepackage{amssymb}
\usepackage{mathtools}

\newcommand{\citep}[1]{\cite{#1}}
\newcommand{\citet}[1]{\citeauthor{#1} [\citeyear{#1}]}

\DeclareMathOperator*{\argmin}{\arg\!\min}
\DeclareMathOperator*{\argmax}{\arg\!\max}

\newtheorem{definition}{Definition}

\newtheorem{example}{Example}
\newtheorem{lemma}{Lemma}
\newtheorem{theorem}{Theorem}

\newtheorem{corollary}[theorem]{Corollary}

\title{Sufficient Decision Proxies for Decision-Focused Learning}
\author{
Noah Schutte$^1$
\and
Grigorii Veviurko$^1$\and
Krzysztof Postek$^2$\And
Neil Yorke-Smith$^1$\\
\affiliations
$^1$Delft University of Technology\\
$^2$Independent Researcher\\
\emails
\{n.j.schutte, n.yorke-smith\}@tudelft.nl,
\{grivurko, krzysztof.postek\}@gmail.com
}

\begin{document}

\maketitle

\begin{abstract}
    When solving optimization problems under uncertainty with contextual data, utilizing machine learning to predict the uncertain parameters' values is a popular and effective approach. Decision-focused learning (DFL) aims at learning a predictive model such that decision quality, instead of prediction accuracy, is maximized. Common practice is to predict a single scenario representing the uncertain parameters, implicitly assuming that there exists a deterministic problem approximation (proxy) that allows for optimal decision-making. The opposite has also been considered, where the underlying distribution is estimated with a parameterized distribution. However, little is known about when either choice is valid. This paper investigates for the first time problem properties that justify using a certain decision proxy. Using this, we present alternative decision proxies for DFL, with little or no compromise on the complexity of the learning task. We show the effectiveness of presented approaches in experiments on continuous and discrete problems, as well as problems with uncertainty in the objective function and in the constraints.
\end{abstract}


\section{Introduction}
\label{sec:intro}
Both in our daily life and in large-scale problems, we often make decisions under uncertainty. Imagine packing your suitcase or finding the best travel route when going to a conference. To make the right decision, we use available data that correlate with what is uncertain: e.g.\@ we use seasonal weather patterns to predict the weather and decide if an umbrella is needed. We are trying to solve a contextual optimization or `predict-then-optimize' problem: making predictions based on contextual information, which we then use for optimization (i.e., decision making). Since our goal is to make the best decision, we do not care about the predictions by themselves, but only about the decision they lead to. This is the main premise of \emph{decision-focused learning} (DFL): learning a predictive model that optimizes for decision quality instead of prediction accuracy.

Defining our setting more formally: Observing contextual information in the form of feature values $z \in Z$, the goal of the decision maker is to find optimal decision $x$ from constrained set $X$ by solving the following problem:
\begin{align} \label{eq:sto}
    x^*(z) = \argmin_{x \in X} \mathbb{E}_{c \sim \mathcal{C}_z}[f(c, x)|z], 
\end{align}
where $\mathcal{C}_z$ is an unknown probability distribution. Since both $c$ and $\mathcal{C}_z$ are unknown, learning a predictive model $\phi_\theta(\cdot)$ for $c$ is a natural approach to the problem, assuming we have data with $z, c$ pairs. After training, once new contextual information $z$ is observed, we can predict an estimated \emph{scenario} $\hat{c} = \phi_\theta(z)$. We do this because we would like to predict the realized value of $c$, i.e., the specific value that $c$ will take in this empirical observation of $z$. If we do this successfully, we will get the empirical optimal decision by solving the \emph{deterministic approximation} (proxy) of the true optimization problem in Equation~\eqref{eq:sto}: $\arg\min_{x \in X} f(\hat{c}, x)$. However, since the true $c$ is stochastic and its distribution $\mathcal{C}_z$ is unknown, the empirical optimal decision is not necessarily the same as the \textit{true} optimal decision based on Equation \eqref{eq:sto}.

In certain settings it has been proven that a single estimated scenario and deterministic proxy allow for optimal decision-making, such as when uncertainty is linear in the objective \citep{Elmachtoub2022}, or for certain fixed recourse, fixed costs two-stage stochastic problems \citep{Homem2024}. However, it has not been studied when a deterministic proxy leads to suboptimal decision-making, and how we can effectively apply DFL in this case. We address this gap in literature by contributing the following:
\begin{enumerate}
    \item We provide sufficient and necessary conditions for when deterministic decision proxies are suboptimal and present problem classes where these conditions are met.
    \item We present alternative decision proxies that adhere to sufficient conditions allowing for optimal decision-making, with little or no compromise on complexity. 
    \item We analyze continuous and discrete problems where deterministic proxies are suboptimal and show the effectiveness of the proposed proxies.
\end{enumerate}

\begin{figure}[ht]
    \centering
\includegraphics[width=\columnwidth]{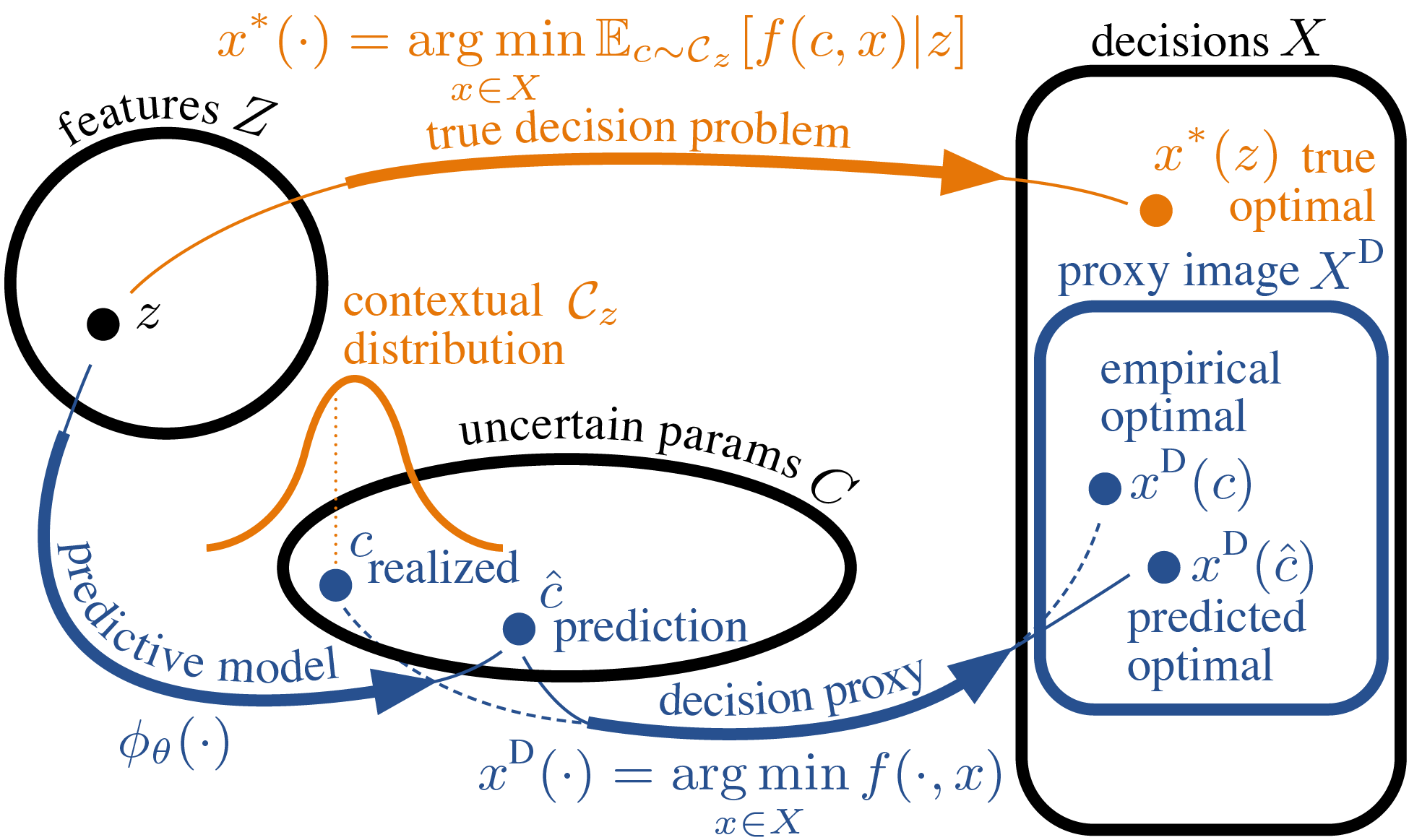}
    \caption{Schematic representation of contextual optimization and the difference between the true optimal decision based on the unknown contextual distribution (orange/top) and the empirical optimal decision based on the deterministic proxy  (blue/bottom).}
    \label{fig:schdfl}
\end{figure}



\section{Problem Formulation}
\label{sec:problem}

The aim in a contextual optimization setting is to go from context $z$ to optimal decision $x^*(z)$, by finding a mapping or policy $\pi: Z \to X$. Due to the structure of the (potentially combinatorial) optimization problem $x^*(\cdot)$, it is challenging to learn this complete end-to-end mapping directly \citep{Bengio2021}. Considering it as separate prediction and optimization problems (predict-then-optimize) is practical, as both problems have been studied extensively \citep{Sadana2025}. From the prediction problem perspective, assuming data with feature-label pairs $z, c$ available, a typical learning would be to train predictor $\phi_\theta: Z \to C$ in a supervised manner by minimizing its predictive error  (e.g., mean squared error: $||c-\hat{c}||^2_2$). We will refer to this approach as \textit{prediction-focused learning} (PFL).  By contrast, DFL aims at learning a predictive model that minimizes decision error, taking the downstream optimization problem into account. Decision error 
is a measure of the difference between empirical optimal decision $x^\text{D}(c)$ and the obtained decision based on the prediction $x^\text{D}(\hat{c})$ (e.g., regret: $f(c, x^\text{D}(\hat{c}))-f(c, x^\text{D}(c))$). Here $x^\text{D}: C \to X$ is the (single scenario) \textit{deterministic proxy} of stochastic optimization problem $x^*(\cdot)$ from 
\eqref{eq:sto}:
\begin{equation*} \label{eq:det}
     x^\text{D}(c) = \argmin_{x \in X} f(c, x). 
\end{equation*}
Both learning methods lead to a policy $\pi^\text{D}= x^\text{D} \circ \phi_\theta$. It is natural to consider this deterministic proxy, since in practice we only have data available and therefore do not have $\mathcal{C}_z$ to compute $x^*(z)$ directly. We observe realized pairs $z, c$, that have an accompanied empirical optimal $x^\text{D}(c)$. However, at inference time, $x^*(z)$ is the optimal decision and there is no guarantee that $x^\text{D}(\cdot)$ is able to equal $x^*(z)$. Proxy $x^\text{D}(\cdot)$ can be limited to a subset of the feasible space $X^\text{D} \subset X$ due to its structure, which means that $x^*(z)$ is unattainable if it lies outside $X^\text{D}$. Figure~\ref{fig:schdfl} shows a schematic representation of this potential problem. We give an illustrative example. 

\begin{example} \label{ex:kelly}
\textbf{Suboptimal investing based on the stock-exchange paradox by \citet{Kibzun1997}.}
Assume we have some capital that we want to grow in the coming $t$ years. Every year, we can deposit capital in a bank with fixed rate of return $\beta=0.05$ and we can purchase shares with an uncertain rate of return $c \sim \mathcal{U}[-1, 1.7]$ (continuous uniformly distributed). If our aim is to maximize the next period's return, we will invest our whole capital in shares, since $\mathbb{E}[c] = 0.35 > 0.05$. However, this leads to a negative \textbf{expected growth rate}:
\begin{align*}
    \mathbb{E}[\ln(1+c)] = \ln(2.7) - 1 < 0,
\end{align*}
Because of this, investing all capital in shares every period will make the probability of ruin, i.e., losing all capital at some point, converge to 1.

This is considered the stock-exchange paradox, where the Kelly strategy \citep{Kelly1956} is being considered as solving:
\begin{align*}
    x^* = \argmax_{0 \leq x \leq 1} \mathbb{E}[\ln(1 + \beta x + c(1-x))].
\end{align*}
Given $c \sim \mathcal{U}[-1, 1.7]$ and $\beta = 0.05$, we get $x^* \approx 0.465$. 

Now, if we look at the deterministic proxy of the problem, it will become clear that we cannot find the optimal decision:
\begin{align*}
    x^\text{D}(\hat{c}) = \argmax_{0 \leq x \leq 1} \ln(1 + \beta x + \hat{c}(1-x)).
\end{align*}
\textbf{Since $\ln(\cdot)$ is a monotonically increasing function, we get:} \textbf{$\boldsymbol{x^\text{D}(\hat{c}) = 1}$ when $\boldsymbol{\hat{c} < \beta} $\quad \& \ $ \boldsymbol{x^\text{D}(\hat{c}) = 0}$ when $\boldsymbol{\hat{c} > \beta}$.} 
Hence true optimal decision $x^*$$\approx 0.465$ cannot be obtained.
\end{example}

\subsection{Contextual Value of the Stochastic Solution}
To more clearly present theory on the suboptimality of deterministic proxies, we extend the concept of \textit{value of the stochastic solution} \citep{Birge1982} to a contextual optimization setting. This concept is introduced as a measure for the value of solving a stochastic optimization problem as compared to solving a deterministic proxy with the uncertain variables replaced by their expectation. This is useful because in practice it is both harder to estimate the uncertain variables distribution (compared to its expectation) and more complex to solve a stochastic optimization problem. Here it gives us some clear structure to prove strict differences between solving a deterministic or stochastic contextual optimization problem. We first define the value of the true optimal decision based on Equation~\eqref{eq:sto}:
\begin{align*}
    \text{V}^* = \min_{x \in X} \mathbb{E}_{c \sim \mathcal{C}_z}[f(c, x)|z] =  \mathbb{E}_{c \sim \mathcal{C}_z} [ f(c, x^*(z)) |z]
\end{align*}

  When we apply PFL, the predictive error is minimized if we are able to exactly estimate $\bar{c}_z = \mathbb{E}_{c \sim \mathcal{C}_z}[c|z]$. Assuming that we can find this estimate, the value of a PFL decision is:
\begin{align*}
    \text{V}_\text{PFL} = \mathbb{E}_{c \sim \mathcal{C}_z}[f(c, x^{\text{D}}(\bar{c}_z))|z].
\end{align*}

In the case of DFL the decision error is minimized. Assuming that we can find $\hat{c}_z$ that minimizes the decision error, the value of the DFL decision is:
\begin{align*}
    \text{V}_\text{DFL} = \min_{\hat{c}_z \in C} \mathbb{E}_{c \sim \mathcal{C}_z} [ f(c, x^{\text{D}}(\hat{c}_z)) |z]
\end{align*}

We have $\text{V}^* \leq  \text{V}_\text{DFL} \leq  \text{V}_\text{PFL}$.  The first inequality holds because in $\text{V}^*$ we directly minimize over the input of $f(c, \cdot)$, $x \in X$, compared to minimizing over the argument of a function with codomain $X$, resulting in the same input of $f(c, \cdot)$. The second inequality holds because we take minimizer $\hat{c}_z \in C$ instead of $\bar{c}_z \in C$. Both inequalities hold for any $z$. This means that if there exists a $z$ with non-zero probability mass for which one of the inequalities is strict, this inequality remains strict in expectation. 

The value of using DFL over PFL $\text{V}_\text{PFL} - \text{V}_\text{DFL}$ is highly relevant for predict-then-optimize problems. A general consensus is that this value increases with predictive error \citep{Mandi2024}, as the higher the predictive error the more important it is to learn such that decision quality is least affected. Additionally, \citet{Cameron2022} show that $\text{V}_\text{PFL} - \text{V}_\text{DFL}$ is also influenced by dependencies between multiple uncertain variables: Predicting marginal expectations using PFL can cause theoretically unbounded worse performance compared to DFL.

In this work we are interested in the difference between the value of using DFL with a deterministic proxy $\text{V}_\text{DFL}$ and true optimal value $\text{V}^*$: $\text{V}_\text{DFL} - \text{V}^*$. 
Due to DFL's end-to-end learning approach, proxy $x^\text{D}(\cdot)$ has not been considered as limiting in its performance, even in more complex stochastic settings \citep{Silvestri2023,Houten2023}. In the following we will identify cases when $\text{V}_\text{DFL} - \text{V}^* > 0$.


\section{Suboptimal Deterministic Decision Proxies} \label{sec:sub}
This section presents cases where DFL with a deterministic proxy is provably suboptimal ($\text{V}_\text{DFL} > \text{V}^*$). We first present theorems for the cases where $\text{V}^* = \text{V}_\text{DFL} = \text{V}_\text{PFL}$ and $\text{V}^* = \text{V}_\text{DFL} < \text{V}_\text{PFL}$. Proofs are presented in the Appendix.

\begin{theorem} \label{the:pfl}
    There exists at least one optimal single-scenario w.r.t.\@ Equation~\eqref{eq:sto} if $\forall z \in Z$, $\mathbb{E}_{c \sim \mathcal{C}_z} [f(c,x)|z] = f(\mathbb{E}_{c \sim \mathcal{C}_z}[c|z], x)$, i.e., Jensen's inequality renders an equality. In this case, an optimal single-scenario is $\bar{c}_z = \mathbb{E}_{c \sim \mathcal{C}_z}[c|z]$. We have $\text{V}^* = \text{V}_\text{DFL} = \text{V}_\text{PFL}$. 
\end{theorem}

The most common application of Theorem~\ref{the:pfl} is when $f(c,x)$ is linear in $c$, $\forall x\in X$. If Jensen's inequality is strict, and hence both functions have different minimizers, PFL becomes provably suboptimal:

\begin{theorem} \label{the:jen}
    If $\argmin_{x \in X}\mathbb{E}_{c \sim \mathcal{C}_z} [f(c,x)|z] \neq \argmin_{x \in X} f(\mathbb{E}_{c \sim \mathcal{C}_z}[c|z], x)$, we have $\text{V}^* < V_{\text{PFL}}$\footnote{Notation is slightly abused here, as the minima's arguments are not necessarily unique. Technically these must be disjoint sets.}.
\end{theorem}

Now we related this to $\text{V}_\text{DFL}$, by presenting a sufficient condition for when a decision proxy allows for optimal decision-making, i.e., an optimal single-scenario exists.

\begin{theorem} \label{the:sur}
    Deterministic proxy $x^\text{D}$ allows for optimal decision-making if $x^\text{D}(\cdot): C \to X$ is surjective, i.e., $\forall x \in X,  \exists \hat{c} \in C: x^\text{D}(\hat{c}) = x$. We have $\text{V}^* = V_{\text{DFL}}$.
\end{theorem}

In other words: When $x^\text{D}$ is surjective, it is expressive enough to attain any feasible decision $x \in X$, which includes optimal $x^*(z)$. Theorem~\ref{the:jen} and Theorem~\ref{the:sur} demonstrate a theoretical case where PFL is inferior to DFL. The result that $\text{V}^* = \text{V}_\text{DFL}$, based on $x^\text{D}(\cdot)$ being surjective, shows the flexibility of DFL: it allows for learning predictions that work well w.r.t.\@ the true optimization problem. On the other hand, if the deterministic proxy is not able to reach stochastically optimal decisions, using this proxy is guaranteed to be suboptimal. This is the main intuition behind suboptimality of the deterministic proxy, and occurs when we have decision dominance: 

\begin{definition} \label{def:dom}
    Decision $x \in X$ is (scenario-wise) dominated if $\forall c \in C, \exists$ (different) $\hat{x} \in X$ s.t. $f(c, \hat{x}) < f(c, x)$. 
\end{definition} 
Note that this definition is based on the objective function $f(\cdot)$, which means that this property does not relate decisions w.r.t. the true stochastic problem in Equation \eqref{eq:sto}. 

\begin{lemma} \label{lem:dom}
    If there exists a scenario-wise dominated $x \in X$, proxy $x^\text{D}(c)=\argmin_{x \in X}f(c,x)$ is not surjective.
\end{lemma}

Lemma~\ref{lem:dom} holds because dominated decisions can not be reached, violating the property of surjectivity. However, non-surjectivity is not a sufficient condition for a proxy to be suboptimal. What is a sufficient and necessary condition is true optimal decision $x^*(z)$ being scenario-wise dominated, as the next theorem expresses.

\begin{theorem} \label{the:dom}
    Proxy $x^\text{D}(\cdot)$ is suboptimal if and only if there exists a feature value $z \in Z$ for which true optimal $x^*(z)$ is scenario-wise dominated. We have $\text{V}^* < \text{V}_\text{DFL}$.
\end{theorem} 

Despite not having access to $x^*(z)$, based on objective function $f$'s characteristics we can prove that dominance occurs for certain continuous decision problems, i.e., when the feasibility set is compact (closed and bounded) and objective function $f$ is continuous. After this we present a class of discrete problems where it is more natural to occur. 

\subsection{Continuous Optimization}
The main result 
for \emph{continuous optimization problems} is the following: Depending on the shape of objective function $f(c,x)$ and expected objective function $g(x):=\mathbb{E}_{c \sim \mathcal{C}_z}[f(c,x)|z]$, it can occur that $\forall c \in C$ the minimizer of $f(c,x)$ lies on the boundary of $X$, while the minimizer of $g(x)$ lies in the interior of $X$. In this case, all interior decisions are dominated by boundary decisions. In Example~\ref{ex:kelly}, we observe that the true objective function is convex non-monotonic in $x$, while the deterministic proxy objective function is strictly monotonic in $x$. Because of this the optimal decision according to the proxy lies on the boundary of the feasibility space, while the true optimal decision lies in the interior. This results in both decisions never coinciding. Example~\ref{ex:kelly} has single-variable functions, but since we are working with multivariate functions in general, we first give a monotonicity definition for multivariate functions.

\begin{definition}
    Let $f: X  \to \mathbb{R}, X \subset \mathbb{R}^m$ be a multivariate function. $f$ is \textit{(strictly) coordinate-wise (non-)monotonic} if $\forall i \in \{1, \dots, m\}$ we have that for any fixed variables $\{\bar{x}_1, \dots, \bar{x}_{i-1}, \bar{x}_{i+1}, \ldots \bar{x}_m\}\in\mathbb{R}^{m-1}$, $f(\bar{x}_1, \dots, x_i, \dots, \bar{x}_m)$ is (strictly) (non-)monotonic in $x_i$.
\end{definition}

Coordinate-wise monotonicity allows the function to be decreasing in some variables while increasing in others.

\begin{theorem} \label{the:subpp}
    Given two continuous (objective) functions $f: C, X \to \mathbb{R}$, $g: X \to \mathbb{R}$, with $g(x):= \mathbb{E}_{c \sim \mathcal{C}_z}[f(c,x)|z]$, and (feasibility set) $X$ is compact, i.e., closed and bounded, and has a nonempty interior. If $f(c,x)$ is strictly coordinate-wise monotonic in $x$ for all $c \in C$, and $g(x)$ is convex and coordinate-wise non-monotonic,  then $\text{V}^* < \text{V}_\text{DFL}$. 
\end{theorem}

In practice the constraints that define $X$ can cause a lower intrinsic dimension and therefore an empty interior. However, in the Appendix we prove that the same result holds.

\begin{example} \textbf{Continuous objective functions with(out) suboptimal deterministic proxy.} (Assuming $X$ is compact.) \\
Objective $f(c,x)=c^Tx$ does not satisfy the conditions of Theorem~\ref{the:subpp}, since $g(x)=\mathbb{E}_{c \sim \mathcal{C}_z}[c^Tx|z] = \mathbb{E}_{c \sim \mathcal{C}_z}[c|z]^Tx$ is linear in $x$ for any distribution $\mathcal{C}_z$ and therefore not coordinate-wise non-monotonic. Similarly $\frac{1}{c^Tx}$ and $(c^Tx)^n$ with even $n$'s do not satisfy the conditions of Theorem \ref{the:subpp} as they are also not coordinate-wise non-monotonic.\\
Objective $f(c,x)=-\log(c^Tx)$ does satisfy the conditions of Theorem~\ref{the:subpp} for at least some distributions $\mathcal{C}_z$. This is because $-\log(c^Tx)$ is strictly coordinate-wise monotonic $\forall c \in C$, because for any $c_i>0$, increasing $x_i$ increases $\log(c^Tx)$, and for any $c_i<0$, decreasing $x_i$ decreases $\log(c^Tx)$. On the other hand, $g(x)$ can be strictly convex with a minimum in the interior, depending on the distribution $\mathcal{C}_z$ (see Example \ref{ex:kelly}). Similarly $\sqrt{c^Tx}, e^{c^Tx}$ and $(cx)^n$ with odd $n$'s satisfy the conditions of Theorem \ref{the:subpp}.
\end{example}

So far we have derived properties for the structure of the optimization problem to show suboptimality. Based on the function examples, one could argue that the types of functions for which this occurs are not common. However, there is an important class of practical optimization problem that typically have complex, non-smooth objective functions.


\subsection{Two-Stage Stochastic Optimization Problems}\label{sec:2stage}
A practical class of optimization problems where a deterministic proxy can be suboptimal is the class of \emph{two-stage stochastic optimization problems}. This two-stage formulation is of particular interest in optimization literature \citep{Birge2011}, as it is effective in modeling real-world problems under uncertainty. These problems consist of making first-stage decisions $x$ that are taken before uncertain parameters $c$ are realized and second-stage decisions $y$ that denote some recourse action after uncertain parameters $c$ are realized. The general formulation is:
\begin{equation} \label{eq:2stg}
\begin{aligned} 
     &\min_{x \in X} f(x) + \mathbb{E}_c [Q(c,x)] \\
    \text{with} \quad &Q(c,x) = \min_{y \in Y(c, x)} g(c,x,y) 
\end{aligned}
\end{equation}

The relevance of this formulation comes from the fact that in practice the realization of uncertainty leads to some action: if we did not bring an umbrella, we can buy a new one.
These \emph{recourse actions} include additional costs compared to what was initially decided. One can model the problem in a way it has \textit{relatively complete recourse}, i.e., the second stage is feasible for any feasible first-stage decision. This is one of the main benefits of two-stage stochastic optimization, as other uncertain problem formulations like (distributionally) robust optimization or chance constraints \citep{Charnes1962} can only ensure feasibility for a specific set of uncertain value realizations or with a certain probability.
Specifically, models with a linear second stage are commonly used: Equation \eqref{eq:2stg} with  $g(c,x,y) = q(c)^T y$ and $Y(c,x)= \{y: T(c)x + U(c)y = h(c) \}$.

\citet{Homem2024} showed that for these problems with fixed costs and fixed recourse, i.e., $q(c)=q$ and $U(c)=U$ being independent of $c$, an optimal single-scenario exists. The proof is based on the assumption that each entry of vector $h(c)$ is dependent on $c$, but when it has independent fixed entries, an optimal single-scenario does not necessarily exist. We demonstrate this with an example based on the weighted-set multi-cover problem in the Appendix. This shows that even for this class of problems it is relevant to consider the used decision proxy. 

\section{Sufficient Decision Proxies} \label{sec:opt}
In the previous section we saw that a sufficient condition for a proxy to be able to lead to an optimal decision is that it is expressive enough. In this section we present proxies that can adhere to this condition, when a deterministic proxy does not. We also discuss why an intuitive choice, using a parameterized distribution proxy, is most often not effective. 
Recall that in a contextual optimization problem, the input to the chosen proxy $x(\cdot)$ is the output of the predictive model $\phi_\theta (\cdot)$, so each proxy asks for a different output from the predictive model. Applying DFL we train this predictive model to get a decision-quality maximizing policy $\pi: Z \to X, \pi = x \circ \phi_\theta$ w.r.t. Equation \eqref{eq:sto}, instead of aiming for the most accurate estimation of uncertain parameters $c$ or distribution $\mathcal{C}_z$.

\subsection{Scenario-Based Proxy}
Based on Theorem~\ref{the:dom} we know that in certain cases proxy $x^\text{D}(\cdot)$ is suboptimal. Since this deterministic variant is introduced to approximate decision problem of interest in Equation~\eqref{eq:sto}, we consider an alternative approximation based on the principle of \textit{sample average approximation} (SAA) \citep{Kleywegt2002}:
\begin{align*} 
     x^{\text{S}}_n(\hat{c}_1, \dots, \hat{c}_n) = \argmin_{x \in X} \sum_{i=1}^n f(\hat{c}_i, x). 
\end{align*}
Note that $x^\text{D}(\cdot) = x_1^\text{S}(\cdot)$. SAA is used to approximately solve stochastic optimization problems, when the uncertain parameter distribution is known but does not lead to a solvable optimization problem (non closed-form objective). This approach works by sampling and subsequently solving the SAA. The SAA converges to the true stochastic optimization problem as the number of samples goes to infinity. In the contextual optimization setting this results in the following theorem:

\begin{theorem}[\protect\citet{Kim2015}]
    Suppose $A \subset X$ is nonempty and compact, such that (i) $x^*(z) \in A$, (ii) for sufficiently large $n$, $x^{\text{S}}_n \in A$ and (iii)  suppose that $f(c, \cdot)$ is continuous at $x$ a.s. for any $x \in X$, and that there exists $\delta > 0$ such that the family of random variables $\{f(c, y): ||y-x|| < \delta\}$ is uniformly integrable. Then $x^{\text{S}}_n(\cdot) \to x^*(z)$.
\end{theorem}

\begin{figure}[ht]
    \centering
    \includegraphics[width=1.0\columnwidth]{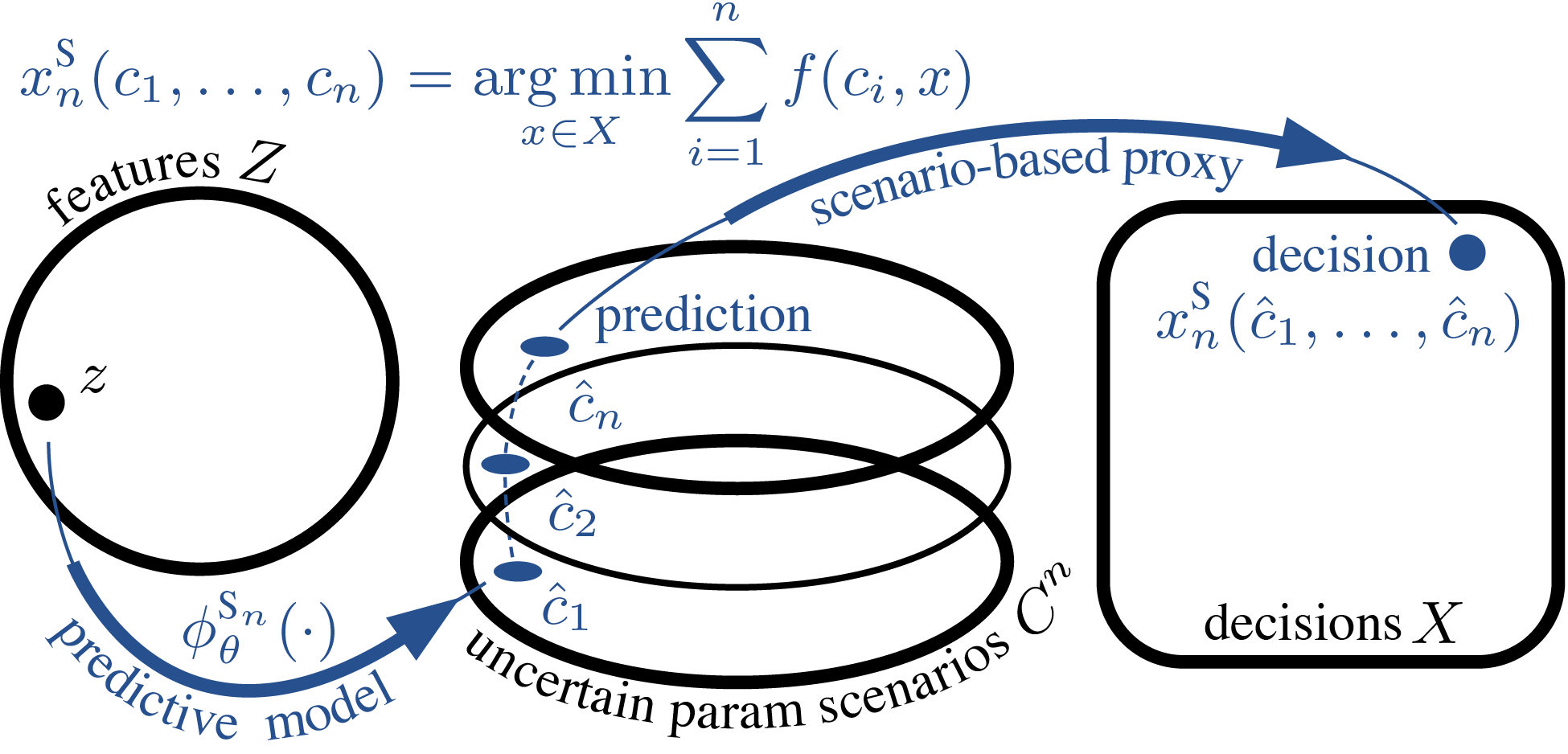} 
    \caption{Scenario-based proxy: Making decisions based on multiple predicted scenarios, given features. }
    \label{fig:sa}
\end{figure}

Based on this theorem, as long as we pick $n$ large enough we can learn to closely approximate the true distribution and get the optimal decision. In a DFL pipeline however, it would be too costly to pick a high value of $n$. Fortunately, the main benefit of DFL is that it can learn to produce high-quality decisions as long as the decision proxy is expressive enough. To achieve this, we predict the scenarios directly instead of sampling them as done in SAA. This results in the following predictor $\phi_{\theta}^{\text{S}_n} : Z \to C^n$, and scenario-based policy $\pi^{\text{S}_n} = x_n^\text{S} \circ \phi_{\theta}^{\text{S}_n}$. In the experimental results we see that with as few as $n=2$ scenarios the scenario-based proxy leads to significantly better decisions than the deterministic proxy. This shows that DFL eliminates the need of using many samples. Figure~\ref{fig:sa} shows a schematic representation of the scenario-based proxy.

\subsection{Quadratic Proxy}
Instead of trying to approximate Equation~\eqref{eq:sto} better, we can also look at the problem from the other perspective: Based on the earlier-developed theory we can design a proxy that has sufficient conditions that allow for optimal decision-making. We can do this by replacing the objective function $f(\cdot)$. First, we generalize the first part of Theorem~\ref{the:sur}:

\begin{theorem}
    Consider a function $x(\cdot): \Xi \to X$. An optimal prediction w.r.t. Equation \eqref{eq:sto} exists if $x(\cdot)$ is surjective, i.e., $\forall x \in X,  \exists \hat{\xi} \in \Xi: x(\hat{\xi}) = x$.
\end{theorem}

\begin{proof}
    Take an arbitrary $z$, with optimal decision $x^*(z)$. Given that $x(\cdot)$ is surjective, $\exists \hat{\xi} \in \Xi$ s.t. $x(\hat{\xi}) = x^*(z)$.
\end{proof}

\begin{figure}[ht]
    \centering
    \includegraphics[width=1.0\columnwidth]{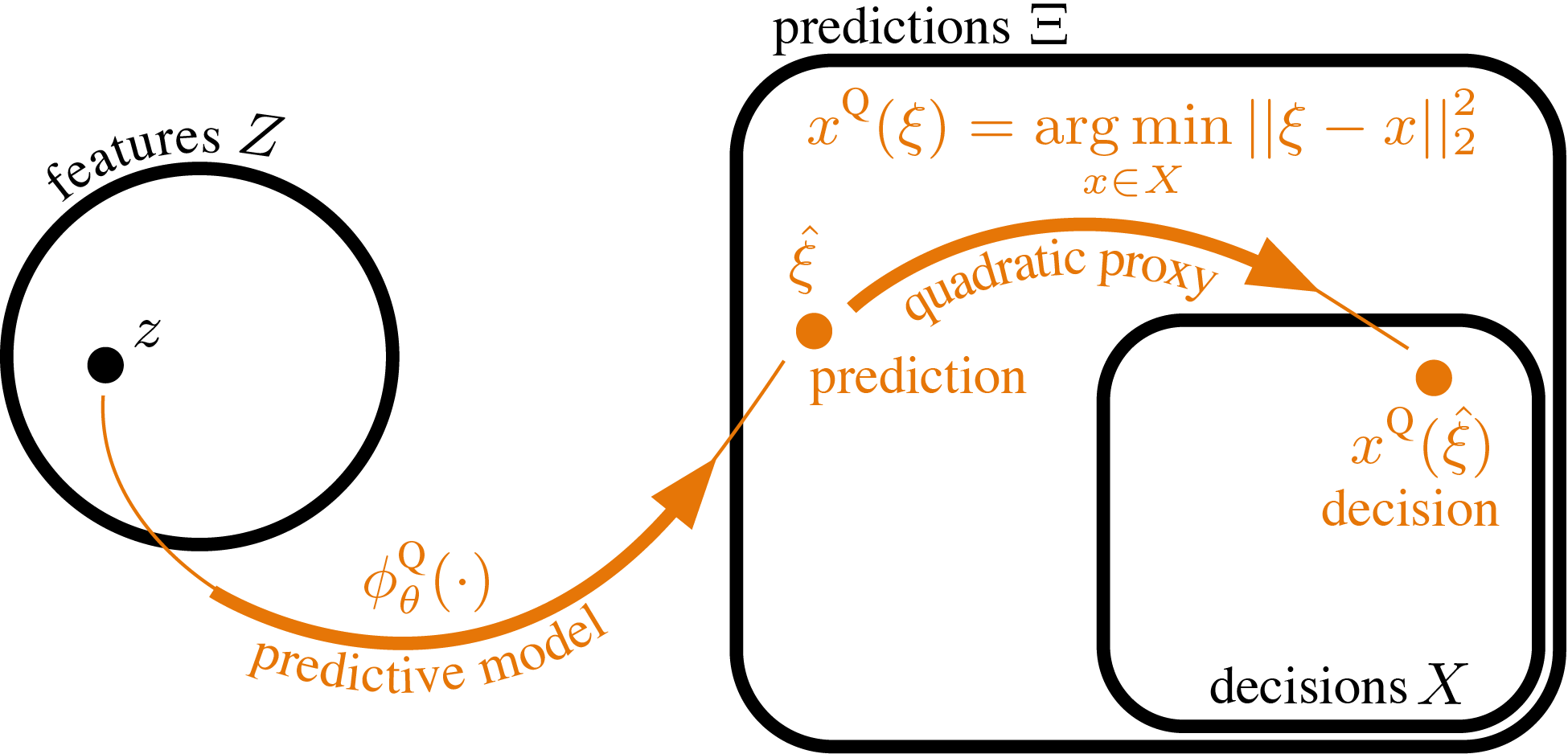} 
    \caption{Quadratic proxy: Making decisions based on predictions in the unconstrained decision space, given features.}
    \label{fig:q}
\end{figure}

We now introduce a surjective function for this setting, which coincides with the \textit{quadratic approximation} proposed by \citet{Veviurko2023} to deal with the zero-gradient problem and is the projection predictions on the decision space.
\begin{align*} \label{eq:q}
     x^{\text{Q}}(\hat{\xi}) = \argmin_{x \in X} ||\hat{\xi}-x||^2_2. 
\end{align*}
This proxy is surjective by design, as for arbitrary $x \in X$, taking $\hat{\xi}=x$ leads to $x^\text{Q}(\hat{\xi})=x$. This leads us to using predictive model $\phi_\theta^\text{Q}: Z \to \Xi$, where $\Xi = \mathbb{R}^{\dim(X)}$, and quadratic policy $\pi^\text{Q} = x^\text{Q} \circ \phi_\theta^\text{Q}$. Figure~\ref{fig:q} shows a schematic representation of the quadratic proxy. 

We elaborate more on the rationale of this proxy $x^{\text{Q}}(\cdot)$: it has a strictly convex objective function, while adhering to the constraints of the problem (feasibility set $X$). This means we can train with DFL and always reach feasible decisions. Based on the assumption of the predictive model being a universal approximator, the objective function does not have to mimic the true objective function in Equation~\eqref{eq:sto}. This is because an optimal decision always lies in some minimum, and this minimum can be approximated by a quadratic function. The DFL pipeline should ensure the predictor learns to predict in the right area. In this case, this area is in the same space as the decisions, where the proxy projects the prediction in the feasibility space. \textit{So effectively we are predicting directly from features to approximate decisions, without considering the uncertain parameters in between.} Most approaches in this category have some correction step to make the decision feasible \citep{Kotary2024}. In this case this can be considered to be the decision proxy itself.  

 \subsection{Parameterized Distribution Proxy}
Since we argue that deterministic proxies and therefore single-scenario predictions can be suboptimal, an intuitive approach would be to predict a distribution and use this to approximate $\mathcal{C}_z$ in Equation \eqref{eq:sto}. This leads to the following decision proxy, with $\beta \in B$ the parameters of the chosen distribution $\mathbb{P}$ and $p_\beta(\cdot)$ its density function:
\begin{align*} 
     x^{\mathbb{P}}(\beta) = \argmin_{x \in X} \int_C f(c,x)p_\beta(c)dc .
\end{align*}
This leads to using predictive model $\phi_\theta^\mathbb{P}: Z \to B$ and policy $\pi^\mathbb{P}=x^\mathbb{P} \circ \phi_\theta^\mathbb{P}$. This approach is discussed as it is effective when the decision proxy has a closed-form, as in one of the experiments by \citet{Donti2017} where the underlying distribution is Gaussian. In practice however, the chosen distribution is unlikely to result in a closed-form integral, especially when problems are more complex, such as the class of problems we aim to tackle. It is still possible to approximate this proxy by discretizing the integral and using SAA. However, this indirect approach of first predicting a distribution and then extracting scenarios is a more restricted version of the scenario-based approach, always resulting in poorer performance. Therefore, we did not include the results.

\subsection{Complexity}
All presented policies have a different combination of predictors and decision proxies. Table \ref{tab:dec} summarizes the policies by denoting the prediction space and the number of decision variables in a two-stage stochastic optimization setting. This number correlates highly with the number of problem constraints, as any constraint involving a second-stage decision is considered for each scenario. In general more decision variables and constraints increase the complexity of the decision proxy. Despite the focus here being on achieving high-quality policies, the complexity of the proxy is highly relevant as it is solved for each data point and epoch in a DFL pipeline.

The prediction space has less impact. The size of this space impacts the number of parameters of the predictive model. Since in general we (can) use neural networks with vastly more parameters than the dimension of the prediction space, this does not change much. What is more of an open question is whether the prediction space influences the complexity of learning a high-quality policy. In some problems $\dim(X)$ is significantly different from $\dim(C)$, which leads to the question if $\pi^\text{Q}$ is harder to learn than $\pi^{\text{S}_n}$. We see small differences in the experiments that are in line with this hypothesis.

\begin{table}[tb]
\centering
\begin{tabular}{ r l r }
\toprule
policy \hspace{3mm} & prediction space & decision var. \# \\
\midrule
  deterministic \hspace{0mm} $\pi^\text{D}$ \hspace{0.5mm} & \hspace{7mm} $C$ & $d_y + d_x $ \hspace{3mm} \\ 
 scenario-based \hspace{0mm} $\pi^{\text{S}_n}$& \hspace{7mm} $C^n$ & $n d_y + d_x$  \hspace{3mm} \\
 quadratic \hspace{0mm} $\pi^{\text{Q}}$ \hspace{0.5mm} & \hspace{7mm} $\mathbb{R}^{\dim(X)}$ & $d_y + d_x$ \hspace{3mm} \\
 distribution \hspace{0mm} $\pi^{\mathbb{P}}$ \hspace{0.5mm} & \hspace{7mm} $B$  &   $d_y + d_x$ \hspace{3mm} \\
\bottomrule
\end{tabular}
\caption{Summary of different policies, the output space of the predictor and the number of decision variables in the decision proxy. $d_x$ and $d_y$ denote the number of first- and second-stage variables.}
\label{tab:dec}
\end{table}


\section{Experimental Evaluation}
\label{sec:exp}
To experimentally evaluate the proposed approaches as alternatives to a deterministic proxy, we consider one continuous and two discrete problems. We use three baselines: deterministic proxy learned using DFL $\pi^\text{D}$, deterministic proxy learned using PFL $\pi^\text{D}_\text{PFL}$ and a PFL learned model using SAA at inference $\pi^\text{S}_\text{PFL}$. In the PFL approaches we train on mean squared error, where $\pi^{\text{S}_n}_{\text{PFL}}$ uses the same learned predictive model as PFL $\phi_\theta$, but evaluates based on an SAA approximation $x^\text{S}_n$. This is done by using a residual-based distribution \citep{Sadana2025}, using empirical additive errors \citep{Deng2022}: taking the predictions as mean, and adding samples from the observed residuals from the validation set. We use $n=16$, so $\pi^{\text{S}_{16}}_{\text{PFL}}$ in all problems. We compare the baselines with our proposed approaches: $\pi^\text{Q}$, $\pi^{\text{S}_n}$, with $n \in \{2, 8, 16\}$. For the continuous problem we apply DFL using differentiable convex layers \citep{Agrawal2019} as the problem is differentiable. For the other problems we use score function gradient estimation  \citep{Silvestri2023}, which is approximates gradients by using stochastic sampling and is therefore applicable to any non-differentiable problem. Regarding the predictors we use neural networks with 2 hidden layers of size 256 and LeakyReLU activation functions, a learning rate of $5 \times 10^{-4}$ and batch size of $32$. Each approach has the same number of epochs, which we consider enough to converge. The best performing model on validation is used for test. Adam optimizer is used for gradient descent and \textit{Gurobi} \citep{Gurobi2020} as the solver. Code is available\footnote{\url{https://github.com/PyDFLT/PyDFLT/tree/sufficient-decision-proxies-ijcai2026}}, based on \textit{PyDFLT} \cite{PyDFLT2026}. For further experimental details please refer to the Appendix. 

\emph{We consider problems where the deterministic proxy is not sufficient.} Since literature so far has omitted these problems, we modified problems from literature to fit this requirement.

\paragraph{Portfolio optimization.}
As considered in earlier work on DFL \citep{Wang2020,Veviurko2023}, we consider a portfolio problem based on data from \citet{Quandl}. Instead of balancing return and risk, we maximize the long-term expected value of the portfolio using the Kelly strategy \citep{Kelly1956}. This makes the problem a multivariate version of Example \ref{ex:kelly}. We determine the fixed rate of return $\beta$ to be the median of all the fourth highest returns per training instance. Since data is limited (2898 data points) we do not use different data splits, instead we randomly pick 10 different securities for each seed, running 10 different seeds. We use a train, validation, test split of 70\%, 15\%, 15\%.

\paragraph{Weighted-set multi-cover.}
We test on the WSMC problem as presented by \citet{Silvestri2023} with one adjustment: We allow removing single-item sets in the case of excess coverage for a partial return of costs, which makes the recourse similar to the common two-stage knapsack problem with uncertain weights \citep{Kosuch2011}. The deterministic proxy can be suboptimal in this case (see Appendix for example). We run 10 random seeds and for train, validation and test set sizes, we use 1000, 250 and 1250.

\begin{figure}[b]

\includegraphics[width=\columnwidth]{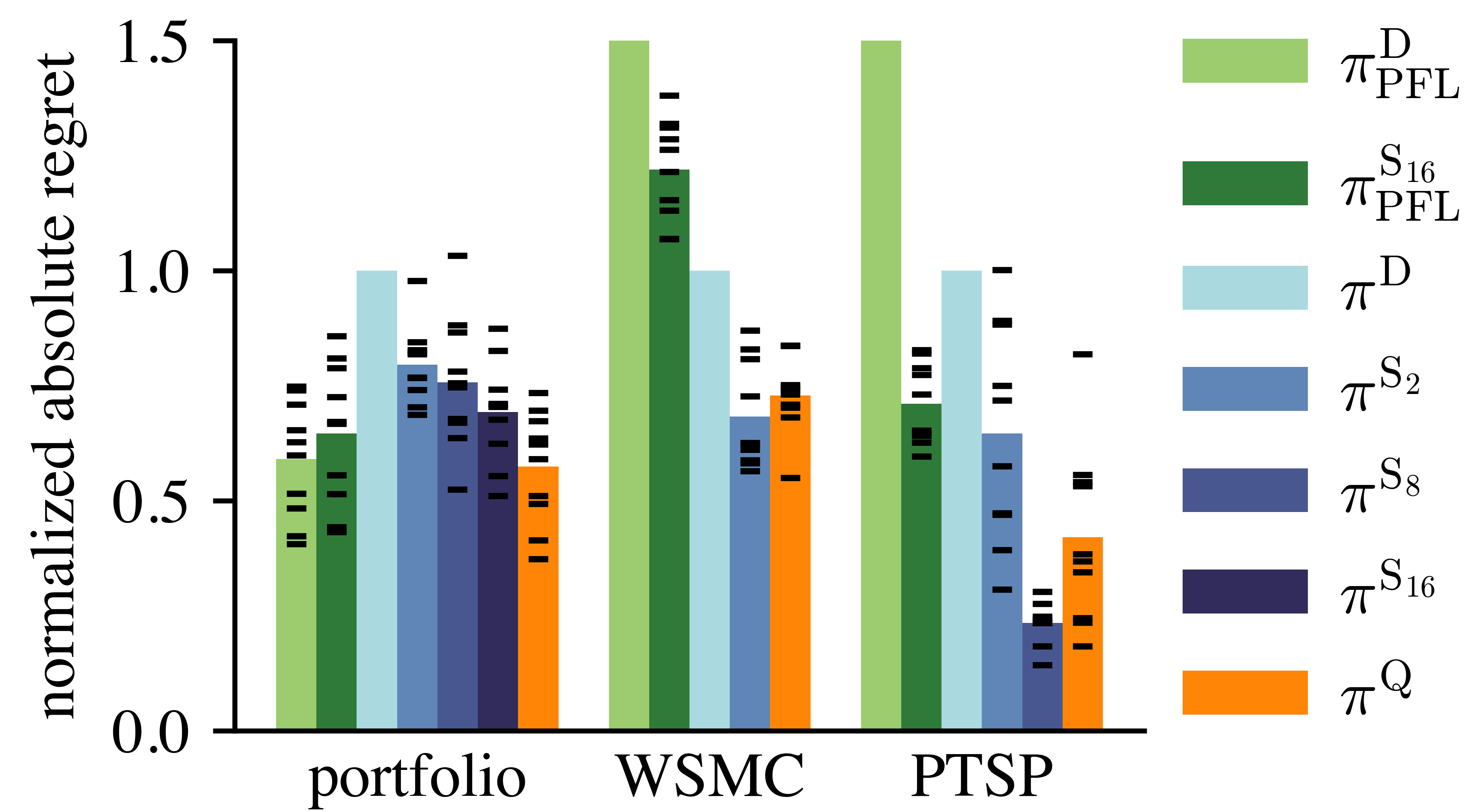} 
    \caption{Test set mean absolute regret normalized by DFL baseline $\pi^\text{D}$. Comparing with PFL baselines and proposed $\pi^{\text{S}_n}$,  $\pi^\text{Q}$. Dashes denote individual runs.  $\pi^\text{D}_\text{PFL}$ bars for WSMC (2.08) and PTSP (2.63) were cut off for visibility. Due to effectiveness of $n=2$ for WSMC, $n=8$ for PTSP, and runtime, higher values of $n$ were not included.}
    \label{fig:test}
\end{figure}

\begin{figure*}[t!]
    \centering
    \subfloat[Portfolio optimization\label{fig:portfolio}]{%
        \includegraphics[width=0.31\textwidth]{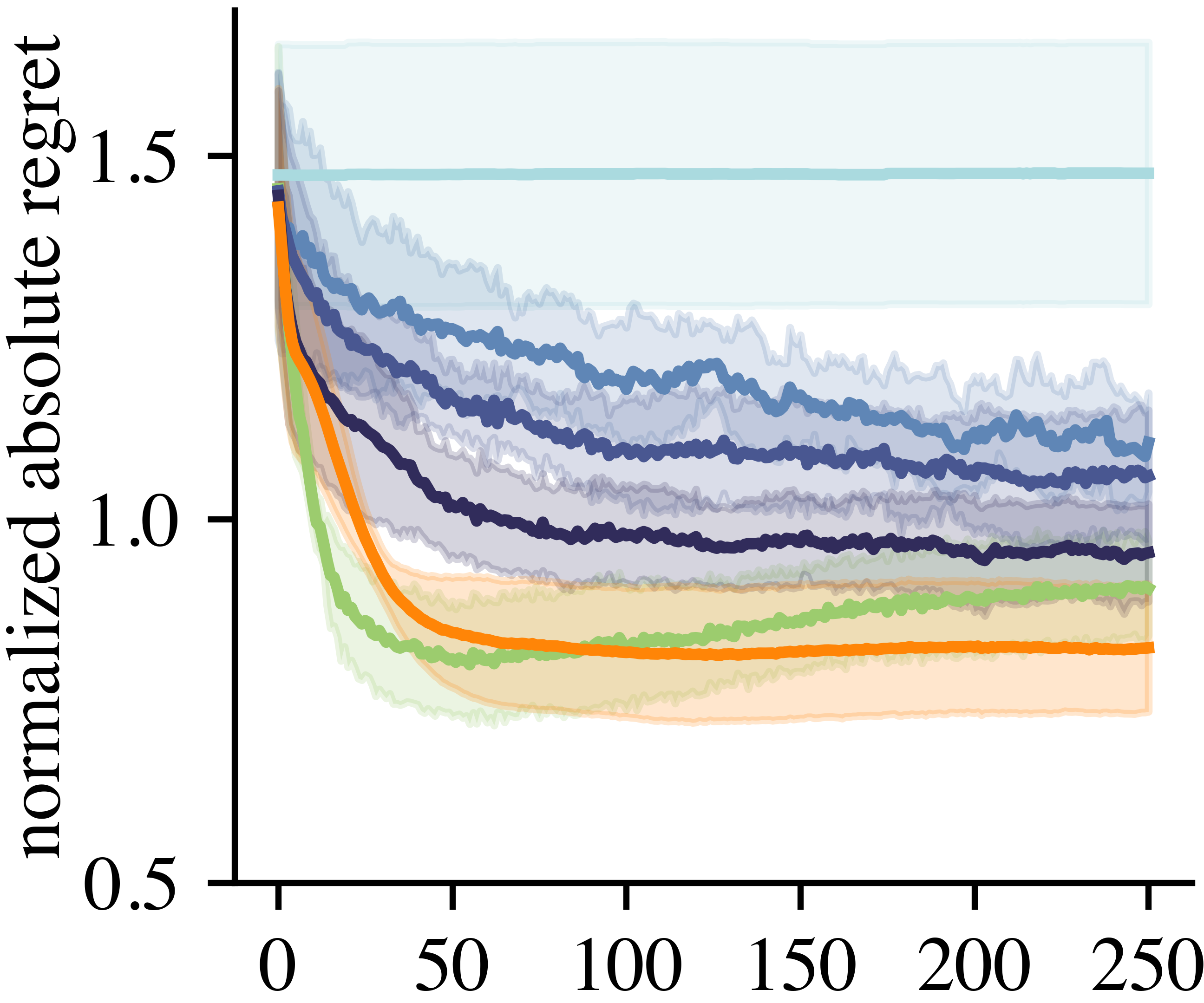}%
    }\hfill
    \subfloat[Weighted-set multi-cover\label{fig:wsmc}]{%
        \includegraphics[width=0.28\textwidth]{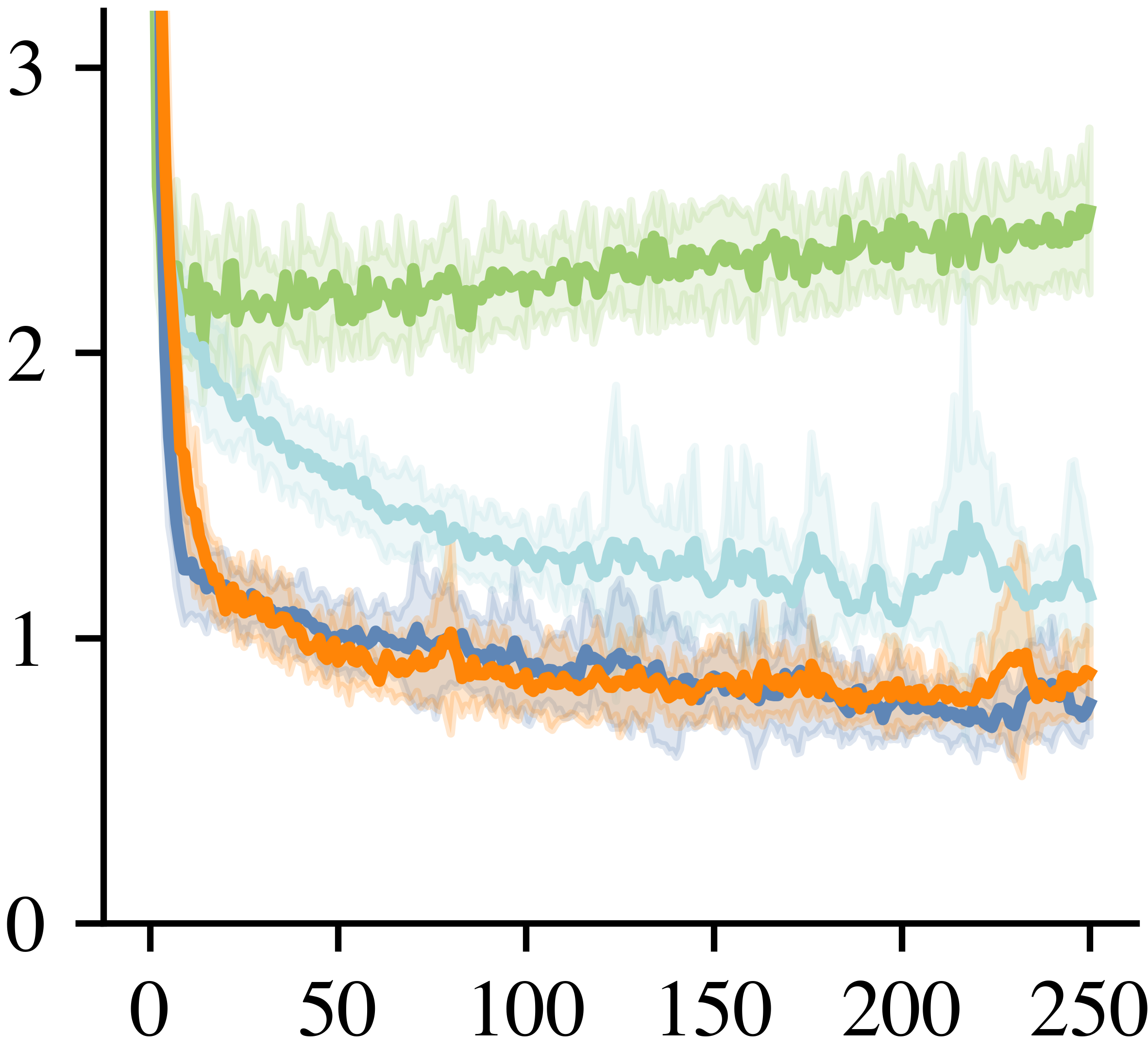}%
    }\hfill
    \subfloat[Probabilistic traveling salesperson\label{fig:ptsp}]{%
        \includegraphics[width=0.28\textwidth]{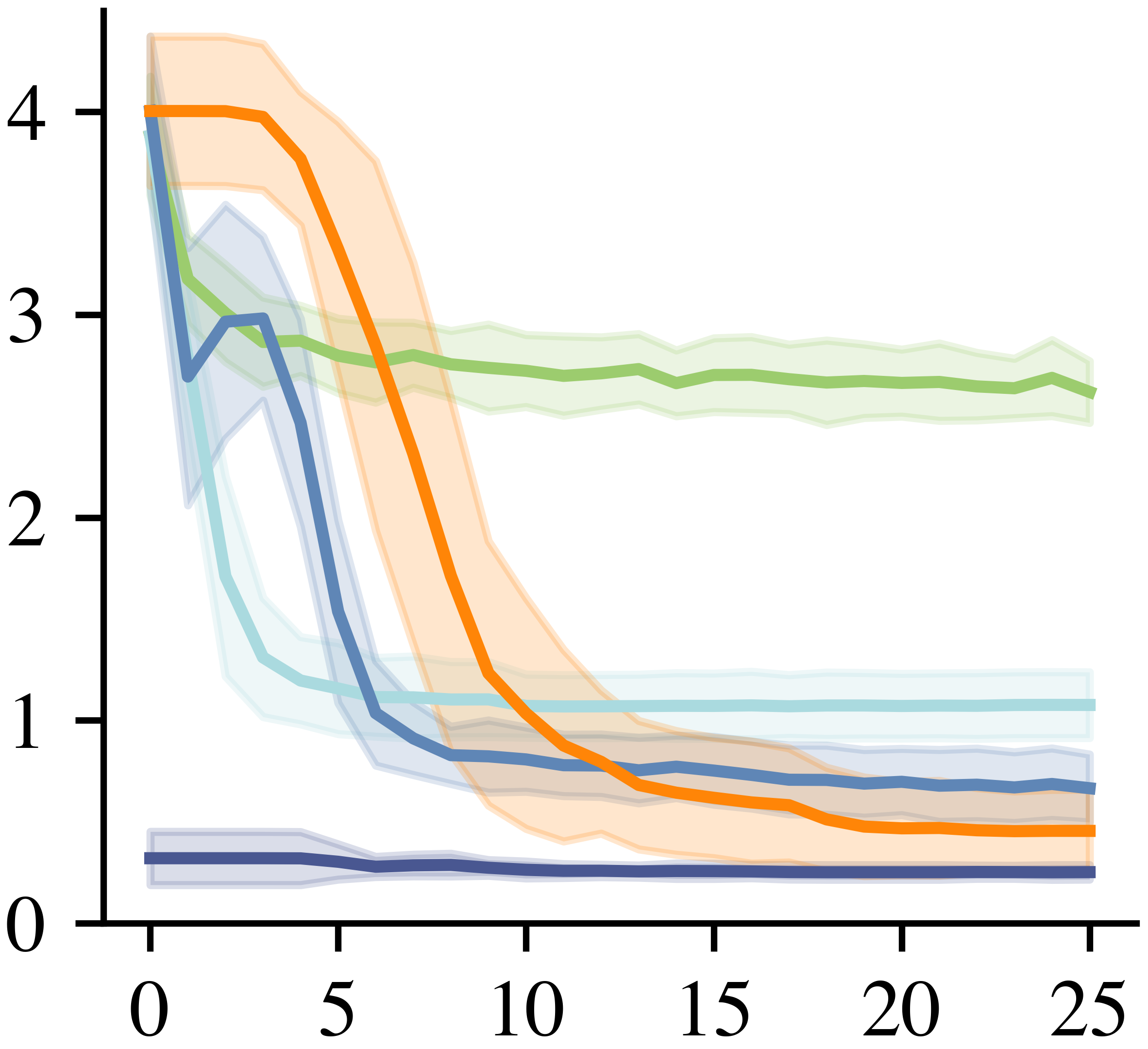}%
    }\hfill
    \subfloat{%
        \includegraphics[width=0.1\textwidth]{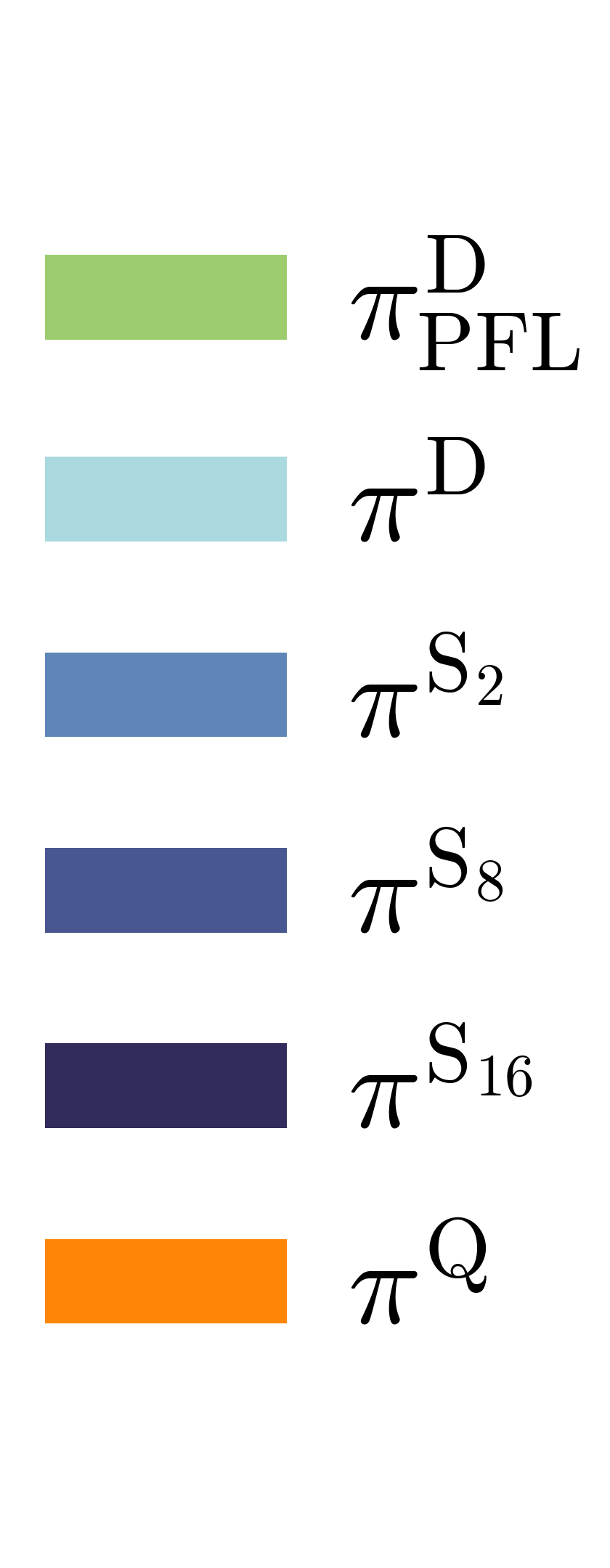}%
    }
    \caption{Validation curves per epoch ($x$-axis). The mean of all approaches' minimum absolute regret was used to normalize. Error bars denote one standard deviation. PFL-approaches do not record absolute regret, but $\pi^{\text{D}}_\text{PFL}$ was included for comparison.}
    \label{fig:val}
\end{figure*}

\paragraph{Probabilistic traveling salesperson.}
Finally we test on a two-stage probabilistic traveling salesperson (PTSP). Probabilistic vehicle routing problems have been studied since \citet{Tillman1969}, with PTSP specifically introduced by \citep{Jaillet1985}, and are still highly relevant \citep{Oyola2018}. The goal of the PTSP is to plan a route along a known set of customers while minimizing transportation costs, but it is uncertain which customers require service. Missing a customer who requires service incurs a penalty (based on going back-and-forth to the depot, which is a typical recourse choice \citep{Oyola2018}). In this two-stage version, one is allowed to add direct trips to customers, similar to allowing crowdsourcing \citet{Santini2022}. After it becomes known which customers require service, these direct trips can be canceled in the second stage if they are not required. This makes the problem such that in a given scenario, direct trips are always dominated, but given uncertainty they have the benefit of incurring low costs if canceled. We run 10 seeds and for train, validation and test set sizes, we use 1000, 250 and 1250.

\subsection{Results}
Figure \ref{fig:test} shows the final results, while Figure \ref{fig:val} visualizes the learning. We use a paired $t$-test (5\% significance level) to compare our proposed methods with the deterministic proxy baseline, and they performed significantly better in all cases. We see that $\pi^\text{Q}$ performs best or second best in all cases. In both WSMC and PTSP the dimension of decision variables is greater than the dimension of uncertain variables. Since this makes $\pi^\text{Q}$ learn the same relationship through a mapping with higher dimension, it could be the reason the scenario-based approach performs slightly better. In general we see that more scenarios for the scenario-based approach results in better performance. For portfolio, Figure \ref{fig:portfolio} shows that $\pi^\text{D}$ is not able to learn at all. This is primarily caused by having zero-gradients. A higher number of scenarios here results in more effective gradients. The most notable part about the WSMC is that the learning curves have high variance. This is inherent to the score function gradient estimation approach. For PTSP this is not as pronounced, due to the small problem size and therefore less complex decision landscape. However, for PTSP we do observe higher variance in the test results for $\pi^\text{Q}$ and $\pi^\text{S}_2$, suggesting problem-dependent lower robustness.


\section{Discussion and Related Work}
\label{sec:discuss}

We formally showed cases when deterministic proxies are suboptimal, even in a DFL setting, and introduce effective alternative proxies. This is another step in understanding the applicability of DFL, after \citet{Cameron2022} consider when PFL is worse compared to DFL and \citet{Homem2024} show for a certain class of problems that an optimal single-scenario exists. Similarly, \citet{Schutte2024} discuss the importance of the decision proxy by highlighting the difference between the empirical (deterministic) problem and the true stochastic problem, and introduce loss functions that generalize better. Furthermore, \citep{Wilder2019} introduced a continuous relaxation proxy to speed up DFL, using a quadratic regularization term on the decision vector.  

There are several works that consider the true problem as stochastic and use distribution-based decision proxies. These works differ from our work in that we provide theoretical backing on when a deterministic proxy is suboptimal, as well as propose alternatives based on this theory that keep complexity low. \citet{Donti2017} tackle a problem that has a differentiable formulation when assuming the uncertainty is Gaussian. Alternatively they assume the uncertain parameter has finite discrete support, having the predictive model output probabilities for these discrete scenarios. This is a similar assumption as done by \citet{Qi2023}, which is quite limiting as uncertain parameters are often continuous and multi-dimensional, requiring to decide on some discretization that is representative without having too many scenarios. Our scenario-based approach learns representative scenarios and only requires to decide the number of scenarios up front. \citet{Elmachtoub2023} also include a distribution-based approach, assuming a parameterized family of distributions. They show that when the distribution is misspecified, distributional DFL (contextual integrated-estimation-optimization) outperforms distributional PFL (estimate-then-optimize), while in a well-specified distribution case the result is opposite. \citet{jia2025} use scenarios in a predict-then-optimize routing problem, but do not train end-to-end.

Stochasticity is also used to obtain gradients when they do no exist or are zero, but in these cases the deterministic proxy is applied at inference \citep{Berthet2020,Silvestri2023}. The same is done at inference by \citet{Hu2023}, who consider uncertainty in the constraints and model infeasibility using penalties. We refer the reader to \citep{Qi2022} for an operations management perspective on DFL, and \citep{Mandi2024} for a general DFL survey.


\section{Conclusions}
\label{sec:conc}

A prevailing assumption in decision-focused learning is that there exists a single-scenario problem approximation that allows for optimal decision-making.  While this is a valid assumption in a wide class of problems, this paper investigates for the first time general theoretical properties of problems for which this assumption is violated. Based on these properties we derive requirements for problem approximations that are sufficient to reach optimal decisions. On three problems we empirically showed the value of the proposed approaches.
Future work includes studying problems where parameterized distributions can easily be assumed and problems from practice, to evaluate the applicability of DFL and different proxies. Given the strong performance of the quadratic proxy, it raises the question how relevant the true objective is in DFL.

\newpage
\clearpage

\section*{Acknowledgments}
We thank the reviewers for their suggestions. This work was partially supported by Epistemic AI and by TAILOR, both funded by EU Horizon 2020 under grants 964505 and 952215 respectively.

\bibliographystyle{named}
\bibliography{library}

\newpage
\clearpage
\appendix

\section{Proofs}
Below we present the proofs for the theory presented in the main work. 
\begin{proof}[Proof for Theorem~\protect\ref{the:pfl}]
\begin{align*}
    \text{V}^* &= \min_{x \in X} \mathbb{E}_{c \sim \mathcal{C}_z}[f(c, x)|z] \\
    &= \min_{x \in X} f(\bar{c}_z,x) \\
     &= f(\bar{c}_z, x^\text{D}(\bar{c}_z)) \\
    &= \mathbb{E}_{c \sim \mathcal{C}_z}[ f(c, x^\text{D}(\bar{c}_z))|z] = \text{V}_\text{PFL} \qedhere
\end{align*}
\end{proof}

\begin{proof}[Proof for Theorem~\protect\ref{the:jen}]
    \begin{align*}
        x^\text{D}(\bar{c}_z) &= \argmin_{x \in X} f(\bar{c}_z, x) \\
        & \neq \argmin_{x \in X} \mathbb{E}_{c \sim \mathcal{C}_z}[f(c, x)|z] = x^*(z)  \\
        \implies \\
         \text{V}^* &= \mathbb{E}_{c \sim \mathcal{C}_z}[f(c, x^*(z))|z] \\
         &< \mathbb{E}_{c \sim \mathcal{C}_z}[f(c, x^\text{D}(\bar{c}_z)|z] = \text{V}_\text{PFL} \qedhere
    \end{align*}
\end{proof}

\begin{proof}[Proof for Theorem~\protect\ref{the:sur}]
    Due to $x^\text{D}(\cdot)$ being surjective, for every $x^*(z)$ there exists $\hat{c}_z \in C$ such that $x^\text{D}(\hat{c}_z) = x^*(z)$. We get:
    \begin{align*}
        \text{V}_\text{DFL}  &= \min_{c_z} \mathbb{E}_{c \sim \mathcal{C}_z}[f(c, x^\text{D}(c_z))|z]  \\
        &\leq \mathbb{E}_{c \sim \mathcal{C}_z}[f(c, x^\text{D}(\hat{c}_z))|z] \\
        &= \mathbb{E}_{c \sim \mathcal{C}_z}[f(c, x^*(z))|z] \\
        &= \min_{x \in X} \mathbb{E}_{c \sim \mathcal{C}_z}[f(c, x)|z] = \text{V}^*
    \end{align*}
    And since we have already shown that $\text{V}^* \leq \text{V}_\text{DFL}$, we get $\text{V}^* = \text{V}_\text{DFL}$.
\end{proof}

\begin{proof}[Proof for Lemma~\protect\ref{lem:dom}]
    Let $x \in X$ be dominated by $\hat{x} \in X$.
    \begin{align*}
        &\min_{x \in X} f(c,x) \leq f(c, \hat{x}) < f(c, x) \quad \forall c \\
        &\implies \\
        &x^\text{D}(c) \neq x \quad \forall c \qedhere
    \end{align*}
\end{proof}

\begin{proof}[Proof for Theorem~\protect\ref{the:dom}]
    Take $z \in Z$ such that $x^*(z)$ is scenario-wise dominated. We have that:
    \begin{align*}
        &\forall c \in C, \exists \hat{x} \in X: \quad f(c, x^*(z)) > f(c, \hat{x}) \geq \min_{x \in X} f(c, x)\\
        &\implies \\
        &x^\text{D}(c) = \argmin_{x \in X} f(c,x)  \neq x^*(z) \quad \forall c\\
        &\implies \\
        &\text{V}^* = \min_{x \in X} \mathbb{E}_{c \sim \mathcal{C}_z}[f(c, x)|z] \\
        &\phantom{\text{V}^*} = \mathbb{E}_{c \sim \mathcal{C}_z}[f(c, x^*(z))|z] \\
        &\phantom{\text{V}^*} < \min_{\hat{c}_z} \mathbb{E}_{c \sim \mathcal{C}_z}[f(c, x^\text{D}(\hat{c}_z))|z]  = \text{V}_\text{DFL}
    \end{align*}
    Where the last inequality comes from the fact that $x^\text{D}(c)$ is unequal to the true minimizer $x^*(z)$ for all $c \in C$, so $\text{V}^* \neq \text{V}_\text{DFL}$. Since we have shown that $\text{V}^* \leq \text{V}_\text{DFL}$, the strict inequality must hold. 
    
    Now for the other direction we assume that we have $z \in Z$ such that $\text{V}^* < \text{V}_\text{DFL}$:
    \begin{align*}
        &\mathbb{E}_{c \sim \mathcal{C}_z}[f(c, x^*(z))|z] < \min_{\hat{c}_z} \mathbb{E}_{c \sim \mathcal{C}_z}[f(c, x^\text{D}(\hat{c}_z))|z] \\
        &\implies \\
        &x^\text{D}(c) = \argmin_{x \in X} f(c,x)  \neq x^*(z) \quad \forall c\\
        &\implies \\
        & f(c, x^\text{D}(c)) < f(c, x^*(z)) \quad \forall c
    \end{align*}
    Where the last step holds because there does not exist a $c \in C$ for which $x^*(z)$ is the minimizer of $f(c,x)$ over $X$. The last inequality shows that $x^*(z)$ is scenario-wise dominated, where for each $c \in C$ the dominating decision is $x^\text{D}(c)$. \qedhere
\end{proof}

\begin{proof}[Proof for Theorem~\protect\ref{the:subpp}]
    Given that for arbitrary $c \in C$, $f(c,x)$ is continuous and strictly coordinate-wise monotone, with $X$ closed and bounded, it attains its minimum $x^\text{D}(c)$ on the boundary of $X$ for all $c \in C$. 
    
    We use proof by contradiction to show that $g(x)$ does not attain its minimum on the boundary. Assume that the minimum of $g(x)$ lies on the boundary: $x^* \in \partial X$ . 

    Because of this boundary assumption, at least one of its coordinates must be on its one-dimensional boundary. Let us say this is coordinate $i$. Since $g(x)$ is  coordinate-wise non-monotonic, we know that $g_i(s):=g(x_1^*, \dots, s, \dots, x^*_m)$ is non-monotonic. Since $g(x)$ is convex, one-dimensional $g_i(s)$ is also convex, and its minimum $s^*$ lies on the interior of (the interval of) $g_i(s)$. This means that $x^*_i \neq s^*$, and $g_i(s^*)< g_i(x^*_i)$. If we now define $x^{s*} :=(x^*_1, \dots,s^*, \dots x^*_m)$ to be the same as $x^*$ except for in the $i$-th coordinate, we get $g(x^{s*})=g_i(s^*)<g_i(x^*_i)=g(x^*)$, which contradicts our assumption that $x^*$ is the minimum.
    
    Because of this, the true minimum cannot lie on the boundary, $x^*(z) \notin \partial X$. Since we also have that $x^\text{D}(c) \in \partial X$ for all $c \in C$, the value at minimum $x^*(z)$ of Equation \ref{eq:sto} cannot be attained by $x^\text{D}(c)$ for any $c$, so:
    \begin{equation*}
    \begin{aligned}
    \overbrace{\mathbb{E}_{c \sim \mathcal{C}_z}[f(c,x^*(z))|z]}^{\text{V}^*} <  \overbrace{\min_{\hat{c}} \mathbb{E}_{c \sim \mathcal{C}_z}[f(c,x^\text{D}(\hat{c}))]}^{\text{V}_\text{DFL} }   
    \end{aligned}
    \end{equation*}
    \end{proof}

In the main work, after presenting Theorem~\ref{the:subpp}, we note that $X$ can have a lower intrinsic dimension and therefore an empty interior due to its constraints. If we can reduce to this lower dimension and we have functions with the same properties, the same result holds.

\begin{corollary}
    If there exists a projection $P \in \mathbb{R}^{k \times m}$, $k<m$ such that $PX$, $f(c, Px)$ and $g(Px)$ have the same properties as $X$, $f(c,x)$ and $g(x)$ in Theorem \ref{the:subpp}, then $\text{V}^* < \text{V}_\text{DFL}$.
\end{corollary}

This corollary simply holds because we obtain a set and two functions with the required properties to apply Theorem~\ref{the:subpp}.

\section{Experimental problem details}
For completeness a mathematical formulation of each of the experimental problems is presented in this section, including the parameters used in the problem experiments.
\subsection{Portfolio optimization} 
For clarity the full problem formulation is shown below, given $n$ securities ($n=10$ is used in the experiments), and $i \in \{1, \dots, n\}$:
\begin{align*}
    &\beta \in \mathbb{R} : &&\text{Net rate of return for the bank} \\
    &c_i \in \mathbb{R} : &&\text{Net rate of return for security $i$} \\
    &x_0 \in [0,1] : &&\text{Percentage investment in the bank} \\
    &x_i \in [0,1] : &&\text{Percentage investment in security $i$} 
\end{align*}
\begin{align*}
    \max_{x_0, \dots, x_n}& \ln (1 + \beta x_0 + \sum_{i=1}^n c_ix_i)  \\
    & \sum_{i=0}^n x_i = 1 \\
    & 0 \leq x_i \leq 1, \quad \forall i \in \{0, \dots, n\}
\end{align*}

\subsection{Weighted-set multi-cover} 
In the WSMC problem, we have some items that have a coverage requirement, which is attained by choosing sets that each cover a subset of these items. The goal is to minimize the total costs of the chosen sets. A stochastic version is presented in \citet{Silvestri2023}, where the coverage requirements are uncertain. When a number of sets is chosen, and the coverage requirements are unmet, an additional cost is paid in the second stage to cover the unmet requirement by adding sets. In the formulation we present here, we allow removing single-item sets in the case of excess coverage. This makes the recourse similar to the common formulation of the two-stage knapsack problem with uncertain weights \citep{Kosuch2011}. Below is the formulation of WSMC with $n$ items and $m$ sets, $m > n$. For each item $i \in \{1, \dots, n\}$, there is a set $j=i$ that only covers that item. We get for $i \in \{1, \dots,n\} \text{ and } j \in \{1, \dots, m\}$:


\begin{align*}
    & x_j \in \mathbb{N} : &&\text{Number of set $j$ chosen} \\
    & c_j \in \mathbb{N}: &&\text{Cost for set $j$} \\
    & \zeta_i \in \mathbb{N}: &&\text{Coverage requirement for item $i$} \\
        & a_{ij} \in \{0, 1\}: &&\text{If item $i$ is covered by set $j$, $A=(a_{ij})$, } \\
        &&& a_{ii} = 1, a_{ji}=0 \text{ for } i,j \in \{1, \dots,n\}, i \neq j \\
    & y^+_i, y^-_i \in \mathbb{N}: &&\text{Unmet/excess coverage item $i$} \\
    & c^+_i, c^-_i \in \mathbb{N}: &&\text{Unmet/excess coverage cost item $i$} 
\end{align*}

\begin{align*}
    &\min_x c^T x + \mathbb{E}_{\zeta} [Q(\zeta, x)]  \\
     Q(\zeta, x) = &\min_{y^+, y^-} c^{+T} y^+ - c^{-T} y^- \\
    & A x + y^+ - y^- = \zeta \\
    & y^-_i \leq x_i, \quad i \in \{1, \dots, n\} 
\end{align*}

Below we give an example of decision dominance as presented in Definition~\ref{def:dom} leading to suboptimal decision-making (Theorem~\ref{the:dom}).

\begin{example} \textbf{Decision dominance in the WSMC.} Given the WSMC problem above, with $n=2$ items, $m=3$ sets, costs $c=(4, 4, 7)$, 
$A = \big[ \begin{smallmatrix} 1&0&1\\ 0&1&1 \end{smallmatrix} \big]$ and recourse costs $c^+ =(7, 7)$, $c^- = (3,3)$. First of all, we note that this problem has complete recourse, i.e., for every first stage decision $x$, there exists a feasible second stage decision. Furthermore, 
we note that the optimal decision to the deterministic proxy of this problem never has recourse, i.e., $y^+ = y^- = 0$, because given $\hat{\zeta}$, we select $x$ such that $Ax = \hat{\zeta}$ and $c^+ > (c_1, c_2) > c^-$. Therefore, it simply reduces to $\min_{x: Ax \geq \hat{\zeta}} c^T x$. Secondly, we note that decision $x = (1, 1, 0)$ is dominated by $\hat{x}=(0,0,1)$, because $\hat{x}$ results in the same exact coverage as $x$, but with lower objective $f(\hat{x})=7<8=f(x)$. However, the optimal decision to the original problem can very well be $x$. Take for example $\zeta$ distributed such that $\mathbb{P}(\zeta = (1,1))=0.5=\mathbb{P}(\zeta = (0,0))$, now let $g(x)=c^T x + \mathbb{E}_\zeta  [Q(\zeta, x)]$, we get $g((1, 1, 0))=5 < g((1,0,0)) = 6  < g((0,0,1)) = 7 = g((0,0,0))$.
\end{example}

In the experiments we use the same setup as in \citep{Silvestri2023}: For the data generation we use as parameters a polynomial degree of 5 and noise width of 0.5. 
We take $n=5$ items, $m=25$ sets, and a penalty of unmet coverage $c^+_i= 5\times\max_{j \leq m} \{a_{ij}c_j\}$. Finally, we set the recovery of excess coverage $c^-_i = 0.8 \times c_i$, where sets $j \in \{1, \dots, n\}$ are the single-item covering sets. 

To construct well-posed problem instances, we use a specific initialization procedure. When the costs and item coverage of sets are randomly initialized, this frequently results in multiple dominated sets. A set is dominated if it has a higher cost than another set while covering only a subset of the items of that other set. These sets are never part of an optimal decision and therefore do not need to be considered, effectively making the problem less complex.

To guarantee that all sets are non-dominated, we first generate all sets to have unique item coverage. Additionally, we apply the following logic: We first generate costs for the single-item covering sets. After this, we iteratively generate a cost for each set covering multiple items based on two bounds:
\begin{itemize}
\item A lower bound: The cost must be strictly greater than the maximum cost of all its proper subsets.
\item An upper bound: The cost must be strictly less than the minimum summed cost of any partition of the set into its existing subsets.
\end{itemize}
The final cost for each set is then calculated as the average of these two bounds, ensuring a well-balanced cost and preventing sets from being only marginally viable.

\subsection{Probabilistic traveling salesperson}
The problem instances of the PTSP are generated by placing the customers equally spaced on a circle (radius of 10) and then perturbed with a normal distribution (standard deviation of 5). Contextual data is generated similar to \citep{Elmachtoub2022}, where we use a polynomial degree of 5. Since we are dealing with binary uncertain parameters, we clip generated values and replace a percentage of values with a Bernoulli distribution. In the experiments, this percentage is set to $50\%$. 
Below is a mathematical formulation for the PTSP with set of nodes $N = \{0, 1, \dots, n\}$ (depot represented by $0$), set of customers $N' = N \setminus \{0\}$. We have for $i, j \in N$, $i' \in N'$:
\begin{align*}
    & x_{ij} \in \{0,1\}: &&\text{When arc $(i, j)$ is traversed} \\
    & x^d_{i'} \in \{0, 1\} &&\text{When there is a direct trip to } i' \\
    & x^v_{i'} \in \{0, 1\}:&&\text{When $i'$ is visited (auxiliary)} \\
    & y_{i'} \in \{0, 1\}: &&\text{If direct trip to $i'$ is canceled} \\
    & d_{ij} \in \mathbb{R}: &&\text{Distance between $i$ and $j$} \\
    & \zeta_{i'} \in \{0, 1\}: &&\text{Customer $i'$ requires service} \\
\end{align*}

\begin{align*}
    &\min_{x, x^d} \sum_{\mathclap{i \in N}}\sum_{j \in N} d_{ij} x_{ij} +\sum_{i \in N'} 2d_{0i}x_i^d + \mathbb{E}_\zeta  [Q(\zeta, x^d, x^v)]  \\
    & \quad \quad x^v_i = x_i^d + \sum_{j \in N, j\neq i} x_{ji}, \quad \forall i \in N' \\
    &\quad \quad  \sum_{j \in N, j \neq i}x_{ij}=\sum_{j \in N, j \neq i}x_{ji}, \quad \forall i \in N \\
        &\quad \quad\sum_{i \in N'}x_{0i} \leq 1 \\
        &\quad \quad\sum_{i \in S}\sum_{j \in S}x_{ij} \leq |S|-1, \\
        & \quad \quad \quad \quad \forall S \subsetneq \{i\in N': x^v_i =1,  x^d_i=0\}\\
    & Q(\zeta,x^d, x^v) =  \min_y \sum_{i \in N'}2 d_{0i} \rho \zeta_i  (1-x_i^v) - \sum_{i \in N'}2 d_{0i} y_i  \\
    & \quad \quad \quad \quad  y_i \leq x^d_i (1-\zeta_i), \quad \forall i \in N' \\
\end{align*}

For the experimental problems we use $n=10$ customers, $\rho = 5$ penalty for missing a customer.

\section{Experimental details}
Below we provide some additional experimental details for the interested reader. See Table~\ref{tab:res} for the results from Figure~\ref{fig:test} in the main work in tabular format. All test results are in mean \emph{absolute regret}:  
\begin{align*}
\frac{1}{m}\sum_{i=1}^mf(c_i, \pi(z_i))-f(c_i, x^\text{D}(c_i)),
\end{align*}
where our training data consists of pairs $(c_i, z_i),$ $ i \in \{1, \dots, m\}$. This is one of the most used metrics in DFL, ever since some of the earlier works in the field introduced it \citep{Donti2017,Elmachtoub2022}, and also used in the portfolio optimization setting \citep{Wang2020}. To get more interpretable results, \citet{Elmachtoub2022} normalize by dividing by the mean of all in-data optimal decisions: $\frac{1}{m}\sum_{i=1}^mf(c_i, x^\text{D}(c_i))$. We normalize in the figures to compare more easily to the main baseline (deterministic proxy) and improve interpretability. 

Another common metric used in DFL is \emph{relative regret} \citep{Mandi2024}:
\begin{align*}
\frac{1}{m}\sum_{i=1}^m \frac{f(c_i, \pi(z_i))-f(c_i, x^\text{D}(c_i))}{f(c_i, x^\text{D}(c_i))}.
\end{align*}
These metrics most often give similar results, which we observed as well in initial experiments.

\subsection{Runtime} 
All experiments were run on a server with fourteen 2.0 GHz CPUs running Ubuntu 24.04 with 32 GB RAM. Code was not parallelized, so CPU utilization corresponded to using a single CPU. Table~\ref{tab:runtime} shows runtimes for different approaches. As expected the PFL approaches take significantly less time training and the scenario-based approach increases training time with more scenarios.

A more interesting finding is that the quadratic proxy $\pi^\text{Q}$ is faster than the deterministic proxy $\pi^\text{D}$ in all cases, even when for the WSMC problem twice the number of samples and therefore solver calls were done (see the subsection on score function gradient estimation). 

Another notable observation is the slower increase in runtime relative to the number of scenarios for the portfolio problem compared to the other two problems. This can be explained by the fact that the WSMC and PTSP problems are two-stage problems, where the number of scenarios is a multiplier of the number of second-stage variables, thus significantly increasing the problem's size.

\subsection{Hyperparameters}
For the scenario-based approach, we use different values for the number of scenarios $n$ based on performance, runtime and consistency. PTSP did not improve with $n>8$. WSMC and portfolio started requiring long runtimes for $n=8$ and $n=32$ respectively (see Table~\ref{tab:runtime}), while showing no signs of significant improvement. For consistency we kept $n \in \{2, 8, 16\}$.

As noted in the main work, hyperparameters such as the predictor architecture, learning rate and batch size were the same for all problems. The number of epochs was kept at 250 for portfolio and WSMC and 25 for PTSP. We did not tune hyperparameters, beyond observing that the chosen hyperparameters led to properly converging learning curves. 

\subsection{Score function gradient estimation}
For two of the experimental problems (WSMC and PTSP) we use score function gradient estimation as proposed by \citet{Silvestri2023}. In this approach stochasticity is used to estimate gradients, making it possible to have the optimization model as a complete black box. We use normal distributions in all cases, initializing sigma to be the standard deviation observed in the training data. The sigma parameter is also updated during back-propagation, but feature independent. Initializing sigma's value too close to zero does not give any gradients, while initializing its value too large will cause a lot of variance and therefore slow convergence. In general we used a single sample per data point, except for the quadratic policy $\pi^\text{Q}$ in the WSMC. Here we used $2$ samples, as this seemed to reduce the variance slightly. Even with double the samples this approach was still faster than the deterministic policy baseline $\pi^\text{D}$ (see Table~\ref{tab:runtime}). 

\subsection{Standardization}
As a best practice, we standardize the input feature values. We also standardize the predictive model outputs, shifting them by the training data uncertain parameter mean and scaling them by the standard deviation. We do the same for the quadratic proxy, where in-data optimal decisions are considered as the output. For the scenario-based approach, we use (in-data) quantiles ($\frac{i}{n+1}, i \in \{1, \dots, n\}$) instead of means for the shift. For the portfolio problem we do not standardize the scenario-based proxy. Particularly $\pi^{\text{S}_2}$ performed worse doing this. Standardization causes the initial predictions to be in a smaller range. When the predicted values are pushed too close together, the distinct scenarios effectively collapse into a single scenario, removing the benefit of having multiple scenarios. For $\pi^{\text{S}_8}$ we did not see a significant difference, but left it out for consistency.

\begin{table}[ht]
    \centering
    \begin{tabular}{llll}
    \toprule
          & portfolio & WSMC & PTSP \\
          \midrule
        $\pi_\text{PFL}^\text{D}$ & 1.75 (0.40) & 579.2 (140.1) & 143.2 (17.5) \\
        $\pi_{\text{PFL}}^{\text{S}_{16}}$ & 1.92 (0.48) & 338.4 (78.2) & 38.8 (4.6) \\
        $\pi^{\text{D}}$ & 2.96 (0.12) & 278.5 (62.9) & 55.5 (9.9) \\
        $\pi^{\text{S}_2}$ & 2.35 (0.29) & \textbf{188.0 (58.5)} & 34.3 (9.6) \\
        $\pi^{\text{S}_8}$ & 2.24 (0.43) & - & \textbf{12.8 (2.4)} \\
        $\pi^{\text{S}_{16}}$ & 2.05 (0.36) & - & - \\
        $\pi^\text{Q}$ & \textbf{1.70 (0.38)} & 202.7 (54.3) & 23.9 (13.6) \\
        \bottomrule
    \end{tabular}
    \caption{Absolute regret mean (standard deviation) on the test set. Best values per problem in \textbf{bold}. Portfolio values are in percentages. First 3 policies are baselines.}
    \label{tab:res}
\end{table}

\begin{table}[ht]
    \centering
    {\small
    \begin{tabular}{lrrrrrr}
        \toprule
         & \multicolumn{2}{c}{portfolio} & \multicolumn{2}{c}{WSMC} & \multicolumn{2}{c}{PTSP} \\
         \cmidrule(lr){2-3} \cmidrule(lr){4-5} \cmidrule(lr){6-7}
         & train & test & train & test & train & test \\
        \midrule
        $\pi_\text{PFL}^\text{D}$ & 12:07 & 5 & 41:10 & 1:58 & 3:42 & 1:32 \\
        $\pi_{\text{PFL}}^{\text{S}_{16}}$ & \texttt{"} \hspace{1.7mm} & 7 & \texttt{"} \hspace{1.7mm} & 13:21 & \texttt{"} \hspace{1.7mm} & 6:27 \\
        $\pi^{\text{D}}$ & 2:18:31 & 5 & 6:48:09 & 2:00 & 36:14 & 1:45 \\
        $\pi^{\text{S}_2}$ & 2:21:27 & 5 & 10:13:55 & 2:55 & 45:12 & 2:13 \\
        $\pi^{\text{S}_8}$ & 2:38:15 & 7 & 29:58:54 & - & 1:29:52 & 4:16 \\
        $\pi^{\text{S}_{16}}$ & 3:18:50 & 10 & - & - & - & - \\
        $\pi^{\text{S}_{32}}$ & 6:52:58 & 32 & - & - & - & - \\
        $\pi^\text{Q}$ & 2:14:09 & 5 & 6:34:26 & 1:03 & 22:13 & 1:05 \\
        \bottomrule
    \end{tabular}
    }
    \caption{Runtime mean in hours, minutes and seconds for train and test (inference) for each approach and problem. $\pi_\text{PFL}^\text{D}$ and $\pi_\text{PFL}^{\text{S}_{16}}$ are trained equally but use a different decision proxy at test time. WSMC $\pi^{\text{S}_8}$ is interpolated from a single unfinished run of 125 epochs.}
    \label{tab:runtime}
\end{table}

\end{document}